\documentclass{article} 
\usepackage{iclr2021_conference,times}

\usepackage{amsmath,amsfonts,bm}

\def\eqref#1{equation~\ref{#1}}

\def\1{\bm{1}}

\def\rva{{\mathbf{a}}}
\def\rvb{{\mathbf{b}}}

\def\rvf{{\mathbf{f}}}

\DeclareMathAlphabet{\mathsfit}{\encodingdefault}{\sfdefault}{m}{sl}
\SetMathAlphabet{\mathsfit}{bold}{\encodingdefault}{\sfdefault}{bx}{n}

\usepackage{hyperref}
\usepackage{url}
\usepackage{booktabs}
\usepackage{graphicx}
\usepackage{enumitem}

\newcommand{\bx}{\mathbf{x}}
\newcommand{\by}{\mathbf{y}}

\newcommand{\EE}{\mathbb{E}}
\newcommand{\cT}{\mathcal{T}}
\newcommand{\cM}{\mathcal{M}}
\newcommand{\cL}{\mathcal{L}}

\newcommand{\std}[1]{{\footnotesize\color{darkgray}$\pm$#1}}

\title{The Effectiveness of Memory Replay in Large Scale Continual Learning}

\author{Yogesh Balaji\thanks{Work done while interning at DeepMind, Mountain View, CA.} \\
University of Maryland, College Park, MD 20742, USA \\
{\small\texttt{yogesh@cs.umd.edu}} 
\AND
Mehrdad Farajtabar, Dong Yin, Alex Mott, Ang Li\thanks{Corresponding author.} \\
DeepMind, Mountain View, CA 94043, USA \\
\small\texttt{\{farajtabar,dongyin,alexmott,anglili\}@google.com} 
}

\iclrfinalcopy 
\begin{document}

\maketitle

\begin{abstract}

We study continual learning in the large scale setting where tasks in the input sequence are not limited to classification, and the outputs can be of high dimension. Among multiple state-of-the-art methods, we found vanilla experience replay (ER) still very competitive in terms of both performance and scalability, despite its simplicity.
However, a degraded performance is observed for ER with small memory. A further visualization of the feature space reveals that the intermediate representation undergoes a distributional drift.
While existing methods usually replay only the input-output pairs, we hypothesize that their regularization effect is inadequate for complex deep models and diverse tasks with small replay buffer size. Following this observation, we propose to replay the activation of the intermediate layers in addition to the input-output pairs. Considering that saving raw activation maps can dramatically increase memory and compute cost, we propose the Compressed Activation Replay technique, where compressed representations of layer activation are saved to the replay buffer. We show that this approach can achieve superior regularization effect while adding negligible memory overhead to replay method. Experiments on both the large-scale Taskonomy benchmark with a diverse set of tasks and standard common datasets (Split-CIFAR and Split-miniImageNet) demonstrate the effectiveness of the proposed method.

\end{abstract}

\section{Introduction}
Humans naturally learn concepts and tasks in a sequential order without degrading performance on the previous ones. Can machines do the same?
This setting, known as \textit{continual learning}, poses serious challenges for the deep learning community.
Deep neural networks often fail to maintain the performance on the previous tasks if they cannot fully access their data, a phenomenon called \textit{catastrophic forgetting}. The underlying cause is the assumption that models learn from \emph{i.i.d.} (independent and identical distributed) data. This assumption is often untrue in the real world where data can come sequentially and their distributions can always be evolving.
The solution to this issue will be a key to a system that allows machines to learn continuously in the real world.


As the machine learning models become deeper and wider while the size of the data gets bigger and the tasks become more diverse, the catastrophic forgetting issue becomes more significant and concerning. In recent years, much work has been proposed to address the forgetting problem. However, existing continual learning methods are usually developed and evaluated on relatively small datasets, often with classification tasks only. The scalability of these methods and their effectiveness in real scenarios with a diverse pool of tasks are still an open question and often being overlooked.

The first question we want to study is: \textit{how different classes of methods work in large scale continual learning with a diverse set of tasks?} We conduct a systematic evaluation of various continual learning algorithms using a comprehensive multi-task benchmark based on the Taskonomy dataset~\citep{zamir2018taskonomy}. This dataset contains not only classification tasks but also regression with high dimensional outputs. Interestingly, we observe in this large scale setting that the vanilla memory replay method outperforms a variety of state-of-the-art approaches, based on a range of evaluation protocols using forgetting reduction, generalization, scalability and inference efficiency.

\begin{figure*}[t!]
\centering
\includegraphics[width=0.9\textwidth]{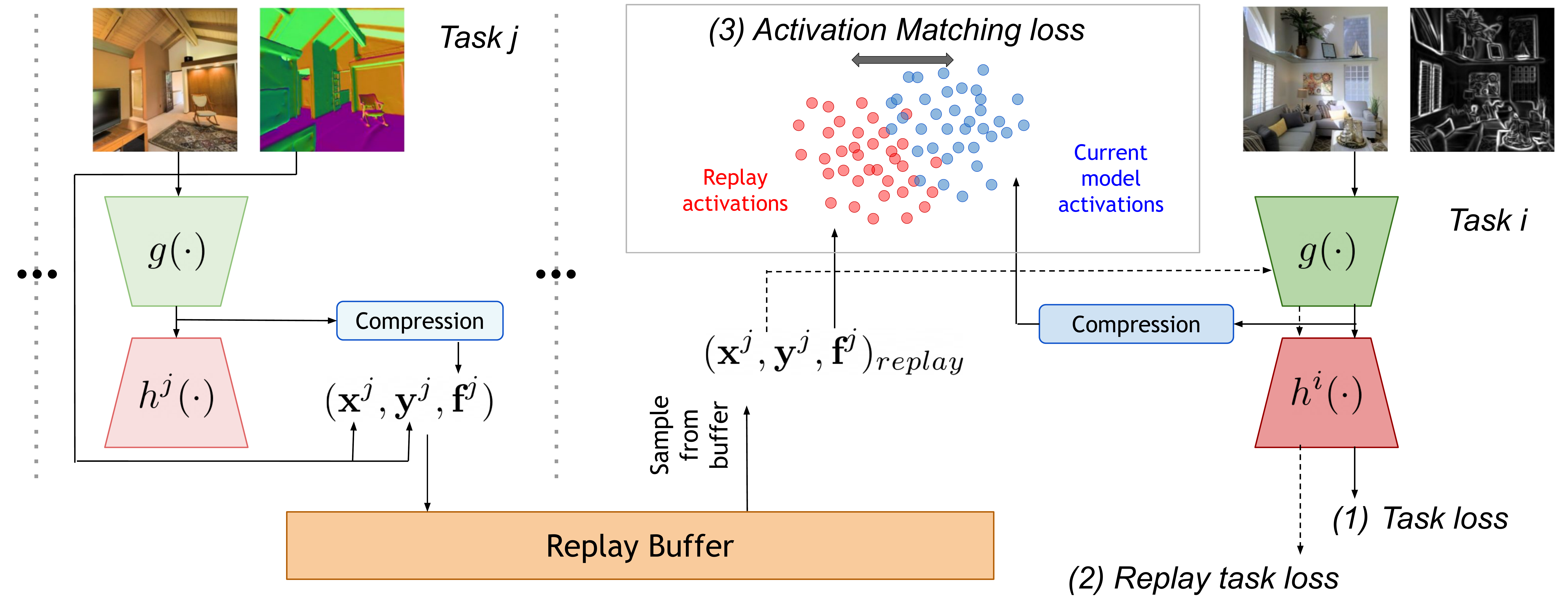}
\caption{\small\textbf{Compressed Activation Replay (CAR)}: During sequential training, compressed feature maps are stored in addition to the input-output pairs in the replay buffer. While training future tasks, models are optimized using (1) task loss, (2) replay task loss using samples from the replay buffer and (3) activation matching loss between the compressed activation maps of the current encoder model and the replay samples. Activation matching loss present distribution drift from happening in the feature space.}
\label{fig:title_fig} 
\end{figure*}


Though being competitive, we observe a degraded performance of vanilla experience replay with a small memory size. A t-SNE plot of the feature space, illustrated in Fig.~\ref{fig:title_fig} top middle, reveals that the intermediate representation still undergoes a distributional shift. We believe it is due to the fact that most replay methods store and regularize using only the input-output pairs from past tasks. Considering that modern neural networks are often wide and deep, we argue that replaying only the input-output pairs are not enough to constrain the feature space. A natural solution is to augment the replay memory with the activation of intermediate layers to regularize representation learning. 


However, given the size of models used in large scale scenarios, these activation tensors can be very expensive in terms of both memory and computation cost. Therefore, we propose a \emph{Compressed Activation Replay (CAR)} framework which regularizes the model by matching the compressed versions of those activation tensors (Fig.~\ref{fig:title_fig}). The chosen compression function is based on average pooling along both spatial and channel dimensions, which results in negligible added compute cost compared to vanilla memory replay. Experimental results also suggest that the compressed replay approach produces equivalent performance compared to replaying the full activation tensors, and outperforms the baseline replay algorithm that only replays the raw input-output pairs.

We evaluate our approach on the large scale Taskonomy benchmark to demonstrate its scalability. New performance evaluation metrics are proposed to tailor the need of large scale setups with diverse tasks. In order to compare with other methods in literature, we also conduct experiments on two medium-sized popular datasets, \textit{i.e.}, Split-CIFAR and Split-miniImageNet. Empirical results suggest that our approach achieves the state of the art in all three benchmarks.

In summary, our contributions are as follows:
\begin{enumerate}
\setlength\itemsep{0em}
    \item We provide a thorough evaluation of popular continual learning methods at a large scale with diverse tasks which is mostly missing in the published literature.
    \item We show the vanilla experience replay is the most effective approach in the large scale setting in the presence of task diversity.
    \item We observe that intermediate representations still undergo a drift in continual learning process reinforcing the catastrophic forgetting. 
    \item We propose Compressed Activation Replay (CAR) to improve the performance of experience replay, reaching state-of-the-art results among replay based methods.
\end{enumerate}

\section{Related work}\label{sec:related_work}
\textbf{Catastrophic forgetting.} Catastrophic forgetting is a major challenge in deep learning where the performance of past examples may drop significantly in a sequential training setting. Continual learning aims to address this problem. The approaches are often categorized into three major directions: replay, expansion and regularization based methods. Replay based methods require a memory module to memorize past or generative examples (see next paragraph for more discussion). Regularization methods are achieved by introducing regularization terms into the loss function which constraint the model weights~\citep{Kirkpatrick_2017,zenke2017continual,ritter2018online,farajtabar2019orthogonal,yin2020sola,Adel2020ContinualLW}. Expansion based methods
\citep{rusu2016progressive,li2017learning,yoon2018lifelong,li2019learn,mirzadeh2020dropout} tackle the problem by expanding the neural architecture along the training. Specifically, the predictive capability on the past examples is retained by freezing the architecture related to past tasks while the new examples are further trained with an extended set of network parameters. These approaches need to expand the neural network, leading to large models in the end. Lacking a proper way for parameter sharing, most of the expansion based methods are found hard to be applied to large scale tasks. 


\textbf{Memory replay methods.}
Mitigating catastrophic forgetting via memory replay is a classic idea that dates back to the 90s~\citep{robins1995catastrophic}. These methods typically save a sampled set of input-output pairs into a memory module and mix these samples with new samples in the current task for training the models.
It has been observed that replay based algorithms usually outperform regularization based methods~\citep{nguyen2017variational,van2019three}, and some explanation have been made by~\cite{knoblauch2020optimal}. Among replay based methods, some try to design new update rules using replay samples~\citep{lopez2017gradient,chaudhry2018efficient}, while others focus on new sampling strategies to fill the replay buffer~\citep{Rebuffi_2017,chaudhry2020hindsightreplay,aljundi2019gradient,Hayes_2019,chrysakis2020online,borsos2020coresets}. Another line of research uses generative models to construct replay examples~\citep{shin2017continual,van2020brain}. Replaying the experience is also popular in reinforcement learning problems \citep{mnih2013playing,schaul2015prioritized,isele2018selective,rolnick2019experience}.

One recent work that is particularly related to our proposed approach is the embedding regularization algorithm~\citep{pomponi2020efficient}. In this algorithm, the network embedding of replay examples are stored in memory and a regularization loss is constructed using the embedding such that the feature extraction layers remain stable. While the basic idea of this work is similar to our proposed method, we differentiate in two ways: 1) in our proposed Compressed Activation Replay method, only a compressed version of the network embedding is stored in the replay buffer, which significantly reduces the memory cost without sacrificing the regularization effect, and 2) we benchmark the performance of the approach on large scale non-classification tasks such as Taskonomy, whereas~\cite{pomponi2020efficient} only provides results based on MNIST and CIFAR.


\textbf{Large scale continual learning.} Most existing continual learning literature still focuses on evaluations on small-scale dataset such as MNIST and CIFAR. The Taskonomy dataset~\citep{zamir2018taskonomy} has been used recently for continual learning study in \citep{sidetuning2020}, where they propose the Side-Tuning algorithm to mitigate catastrophic forgetting. Side-Tuning is an expansion based method, which trains a separate side network for every new tasks and blend the output of the side network with that of the previous frozen networks. The method is able to achieve zero forgetting since all previous tasks' models are frozen. A potential concern for this method is that the inference time of later tasks is significantly higher than that of earlier tasks. Our proposed approach differs from Side-Tuning in that we allow the representation network to be shared and evolved over tasks with stable model capacity and inference time.
Another large scale dataset that is often used in continual learning research is the CORe50 dataset~\citep{lomonaco2017core50}, which mainly focuses on object detection. In this work, we choose Taskonomy since it contains more diverse tasks.
\vspace{-2mm}
\section{Preliminaries}
\vspace{-2mm}
In this section, we introduce the problem formulation and evaluation metrics of the type of continual learning problems considered in this work, together with a review of the memory replay basics.

\textbf{Continual learning.} In continual learning, we aim to solve a sequence of supervised learning tasks $(\cT^1, \cT^2, \hdots \cT^t)$, where, $\cT^{i} = \{ (\bx^i_j, \by^i_j) \}_{j=1}^{N_i}$ denotes the data examples for task $i$.  The objective is to perform sequential training while preserving the performance on the past tasks without access to all of their data.
Let $\ell^{i}(\cdot, \cdot)$ denote the loss function for task $\cT^i$, which can be any generic loss function (\textit{e.g.,} mean-squared-error or cross entropy), and is not tied to any specific task such as classification. In this work, we adopt an encode-decode model. More specifically, let $g(\cdot)$ denote the shared encoder model, and $h^{i}(\cdot)$ the task head for the task $\cT^i$. Predictions on the task $\cT^i$ are made using the model $p^{i}(\bx) = h^{i}(g(\bx))$. The task loss for optimizing  task $\cT^i$ is then given by
\begin{align}
    \cL^{i}_{task} = \EE_{(\bx, \by) \in \cT^i}~~ \ell^{i}(p^{i}(\bx), \by).
\end{align}

\textbf{Memory replay.} In the simplest version of memory replay based continual learning, we maintain a replay buffer $\cM$ of size $m$ per task (total buffer size is $mt$) for storing samples from the task sequence seen thus far. In addition to task loss, models are optimized using a replay loss given by
\begin{align}
    \cL^{j}_{replay} = \EE_{(\bx_{r}^{j}, \by_{r}^{j}) \in \cM}~~ \ell^{j}(p^{j}(\bx_{r}^{j}), \by_{r}^{j}).
\end{align}
Here, $(\bx_{r}^{j}, \by_{r}^{j})$ denotes the samples from the replay buffer belonging to task $j$. In this work, the data in the replay buffer are sampled uniformly at random from the training data for each task. The overall optimization objective in vanilla experience replay (ER) for training task $i$ is then given by
\begin{align}\label{eq:vanilla_er}
    \min_{g, h^i}~~ \cL^{i}_{task} + \frac{\lambda}{i-1} \sum_{j=1}^{i-1} \cL^{j}_{replay},
\end{align}
where, $\lambda$ is a hyper-parameter that balances the current task loss and the replay losses.

\textbf{Evaluation metric.}
One of the challenges in continual learning with non-classification tasks (textit{e.g.,}, Taskonomy dataset) is the evaluation metric. In this work, for Taskonomy, we propose \emph{forgetting} and \emph{performance drop} as two major metrics. Let $\ell_{test}(i, j)$ denote the test loss of task $j$ using the model snapshot obtained at the end of training task $i$ in the task sequence, and let $\ell^{MT}_{test}(i)$ denote the test loss of task $i$ obtained from a multi-task model trained simultaneously on all tasks. Then, the two metrics are defined as follows:
\begin{align*}
    \text{Forgetting :=}&~~~~ \frac{1}{t}\sum_{i=1}^{t} \frac{\ell_{test}(t, i) - \ell_{test}(i, i)}{\ell_{test}(i, i)} \times 100\%     \\
    \text{Performance drop :=}&~~~~ \frac{1}{t}\sum_{i=1}^{t} \frac{\ell_{test}(t, i) - \ell^{MT}_{test}(i)}{\ell^{MT}_{test}(i)} \times 100\%
\end{align*}
The \emph{forgetting} measure indicates the average loss increase of the model compared to the model snapshot at the end of $i^{th}$ task, while the \emph{performance drop} quantifies the performance drop of the final model compared to a multi-task model trained using the same model architecture.

For classification tasks, \textit{e.g.,} Split-CIFAR and Split-miniImageNet, we report \emph{forgetting} and \emph{average accuracy} as two main metrics. Specifically, let $Acc(i, j)$ denotes the accuracy on task $j$ using the model snapshot obtained at the end of training task $i$. We have
\begin{align*}
   \text{Forgetting :=}~~~~ \frac{1}{t}\sum_{i=1}^{t} \max_{j} \left[ Acc(j, i) - Acc(t, i) \right], \quad \text{Avg accuracy :=}~~~~ \frac{1}{t}\sum_{i=1}^{t} Acc(t, i).
\end{align*}

\section{Vanilla experience replay is a strong baseline}\label{sec:vanilla_er}

We begin our study by investigating the performance of different classes of continual learning algorithms on Taskonomy, which is a large scale dataset with diverse set of tasks. We compare to Side-Tuning~\citep{sidetuning2020}, an expansion based method mentioned in Section~\ref{sec:related_work}, EWC~\citep{Kirkpatrick_2017}, a regularization algorithm that constructs the regularizer based on the Fisher information of the model's weights, A-GEM~\citep{chaudhry2018efficient} a memory replay baseline that uses replay examples to construct a constrained optimization problem, and finaly ER the vanilla experience replay algorithm described in~\eqref{eq:vanilla_er}. For completeness, we also compare with the vanilla SGD algorithm that trains the model sequentially on all the tasks without storing extra information, and the multi-task (MT) algorithm which has access to all the tasks simultaneously.


\subsection{Experiment setup}

Taskonomy \citep{zamir2018taskonomy} is a large scale dataset of indoor scenes from various buildings containing annotations for several diverse computer vision tasks. We use data from the following $11$ tasks in this work: \textit{curvature estimation, object classification, scene classification, surface normal estimation, semantic segmentation, depth estimation, reshading, edge detection (texture), edge detection (occlusion), 2D keypoint estimation and 3D keypoint estimation}. All input samples are of $256 \times 256$ resolution. We use a subset of Taskonomy dataset containing $50$ buildings in all our experiments. Of these, $5$ randomly chosen buildings are treated as held-out test set, while the rest of the data are used for training and validation. This yields a total training size of $324,864$ samples.

\textbf{Loss function.} Following \cite{zamir2018taskonomy}, object and scene classification tasks are trained using cross entropy loss, semantic segmentation using pixelwise cross entropy, curvature estimation using $\ell_1$ loss, and all other tasks using $\ell_2$ loss.

\textbf{Model.} We use Resnet-50 model as our encoder network $g(\cdot)$. Since most tasks in Taskonomy dataset are pixel-wise tasks, we do not perform the spatial average pooling in the pen-ultimate layer of Resnet-50 model. This results in a feature map of dimension $2048 \times 8 \times 8$. Following \cite{zamir2018taskonomy}, we use $15$-layer ConvNet model with upsampling blocks as the task heads $h^i$.

\begin{table*}
\caption{\small\textbf{Taskonomy results} using various methods. Forgetting and performance drop ($\Delta$Performance) values are reported. $11$ different tasks are used. All results are averaged over $3$ random seeds. ER-$m$ means we store $m$ data examples for each task. Top two results for forgetting and performance are highlighted in \textbf{bold}. }\vskip -1em
\label{tab:taskonomy_1}
\begin{center}
\resizebox{\textwidth}{!}{
\begin{tabular}{r | cccc  ccc}
\toprule
Method & MT & SGD & Side-Tuning & EWC & A-GEM & ER-64 & ER-256\\
& - & - & (expansion) & (regularization) & (replay) & (replay) & (replay)\\
\midrule
Extra Storage (GB) & 410\footnotemark & 0 & 0.46 & 5.40\footnotemark & 0.44 & 0.11 & 0.44\\
Final model size (GB) & 1.30 & 1.30 & 1.76 & 1.30 & 1.30 &  1.30 &  1.30 \\
Forgetting (\%)& N/A & 191.61 & \bf 0 & 169.62 & 38.88 & 40.18 & \bf 14.73\\
$\Delta$Performance (\%)& \bf 0 & 101.56 & 13.12 & 96.75 & 49.01 & 20.70 & \bf 8.52\\
\bottomrule
\end{tabular}}
\vspace{-2mm}
\end{center}
\end{table*}

\subsection{Results}
Table~\ref{tab:taskonomy_1} contains the result of our large scale analysis. Interestingly, we found that despite its simplicity, the vanilla experience replay algorithm is a strong baseline on Taskonomy. Similar findings were also reported by~\cite{chaudhry2019tiny} but on conventional and mainstream CL datasets like split CIFAR, mini-ImageNet, and CUB which are less realistic, less diverse and smaller in the scale. 
We believe these results indicate that memory replay based algorithm can be a very promising direction for large scale CL that should not be ignored.

We note that in Table~\ref{tab:taskonomy_1} that expansion based method Side-Tuning achieves zero forgetting and competitive performance drop. However, it requires a growing neural architecture which results in a large final model with more expensive inference cost. Its compute overhead will further increase when seeing larger numbers of tasks. After each task, Side-Tuning needs to add an additional side network which is also found expensive in the training storage cost. That prevents this method being applied to complex tasks that requires models with large capacity.

We also observe a degraded performance of vanilla experience replay with a small memory size, \textit{i.e.,} $m=64$ is worse than $m=256$. Although not surprising, the degraded performance inspired our study on the feature space which leads to the proposed improvement approach in the next section.

\footnotetext{This is the estimated training data size excluding the current task which needs access by multi-task learning.}
\footnotetext{EWC needs access to all previous tasks' encoder weights with the diagonal of fisher information.}

\vspace{-2mm}
\section{Compressed Activation Replay}
\begin{figure*}[t!]
\centering
\includegraphics[width=0.9\textwidth]{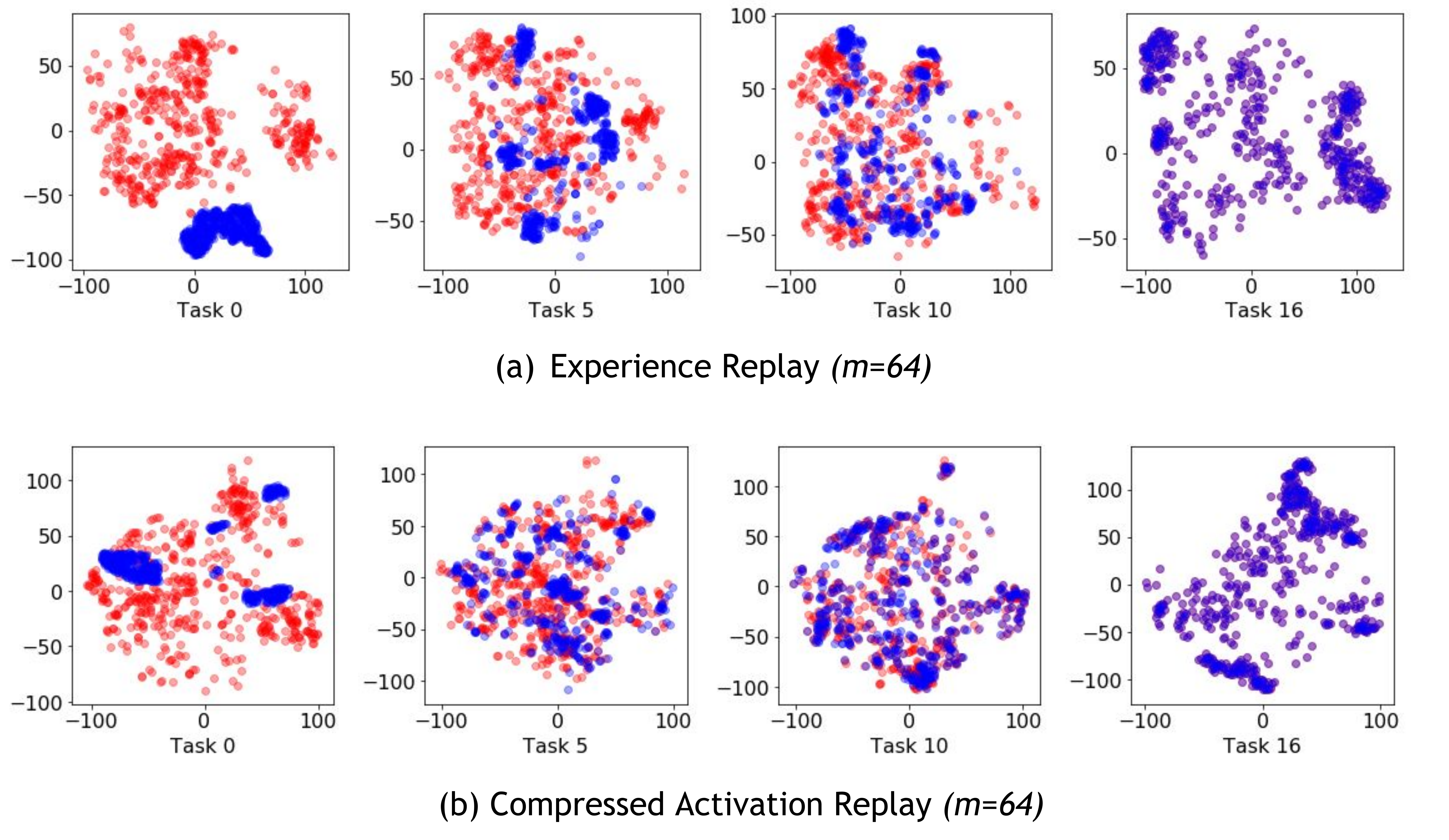}
\vspace{-2mm}
\caption{\small\textbf{Visualizing t-SNE embeddings} of model trained using vanilla experience replay (a) and Compressed Activation Replay (b) on Split-CIFAR dataset. We compare the features of the encoder of the final model (\emph{shown in blue}) relative to the encoder representations obtained while training the respective tasks (\emph{shown in red}). We observe that older tasks encounter significant drift in the feature space when trained with experience replay. Using activation matching loss, our Compressed Activation Replay framework reduces the drift as the blue dots better cover the support of red dots, especially on Task 5 and 10. We note that for the very early task (Task 0), feature drift can still be observed even if we use CAR.
}
\vspace{-4mm}
\label{fig:tsne_ER} 
\end{figure*}

\vspace{-2mm}
Vanilla experience replay stores only input-output pairs and uses them to construct the replay loss. However, we suspect that, for large neural networks, intermediate features (activations) may undergo drifts when we train the model sequentially on different tasks. If that is the case, the input-output pairs may not be sufficient to mitigate this drift in the intermediate layers, leading to catastrophic forgetting. To better understand this phenomenon, we plot the t-SNE  \citep{tsne} visualization of the feature space over the course of vanilla ER in Figure~\ref{fig:tsne_ER}(a). We observe that indeed as new tasks are being learnt, feature representations of the older tasks start drifting. To circumvent this issue, we propose using feature matching as an additional regularization term.



Let $(\bx_r^j, \by_r^j)$ be a sample from task $j$, and let $\rvf_r^j = g(\bx_r^j)$ be the encoder representation (\textit{i.e.,} activation) of this sample at the end of the training of task $j$. The neural net prediction of this sample is thus $p^j(\bx_r^j) = h^j(\rvf_r^j)$. The basic idea of feature matching regularization is that, when we train a subsequent task, say task $i$, the encoder representation of the replay example $\bx_r^j$, \textit{i.e.,} $g(\bx_r^j)$ should stay close to $\rvf_r^j$. To achieve this, we add an additional regularization term $\ell_{fm}(g(\bx_r^j), \rvf_r^j)$ when training the subsequent tasks, where $\ell_{fm}$ is a loss function that measures the difference between $g(\bx_r^j)$ and $\rvf_r^j$. In the following, for most experiments we use a simple MSE loss, \textit{i.e.,} $\ell_{fm} (\rva, \rvb) \equiv \|\rva - \rvb\|_2^2$. In our ablation study in Section~\ref{sec:ablation}, we compare different choices of $\ell_{fm}$.


The activation tensors of intermediate layers in neural networks are usually of high dimensions, and storing them in the replay buffer can be expensive in memory cost. Since we also want to avoid too much overhead to the memory and compute cost, we empirically study a variant of feature matching regularization by compressing the activation into small-length vectors and only regularize the models in regards to such compressed vector representations. Formally, we use a compression function $c$ to map the high dimensional feature $\rvf_r^j$ to a low dimensional vector $c(\rvf_r^j)$, and then apply the feature matching loss $\ell_{fm}$. In this case, the feature matching loss for task $j$ can be written as
\begin{align}
    \cL^{j}_{fm} = \EE_{(\bx_{r}^{j}, \by_{r}^{j}, \rvf_{r}^{j}) \in \cM}~~ \ell_{fm}(c(g(\bx_{r}^j)), c(\rvf_{r}^j)),
\end{align}
and then the objective for training a subsequent task $i$ can then be written as 
\begin{align}
    \min_{g, h^i}~~ \cL^{i}_{task} + \frac{1}{i-1} \sum_{j=1}^{i-1} \left[ \lambda\cL^{j}_{replay} + \lambda_{fm} \cL^{j}_{fm} \right],
\end{align}
where $\lambda_{fm}$ is another coefficient balancing the feature matching loss and other terms in the loss function.
We name this approach Compressed Activation Replay (CAR). As presented in Figure~\ref{fig:tsne_ER}(b), by using CAR, the drift in the encoder representations is significantly reduced.
As for the compression function, a simple way to compress is to perform average pooling along axes (spatial and/or channel) of the tensors. Interestingly, we found this simple compression performs well in our experiments. In our ablation studies in Section~\ref{sec:ablation}, we compare different compression techniques.


\vspace{-2mm}
\section{Empirical studies}
\begin{figure*}[t!]
\centering
\includegraphics[width=\textwidth]{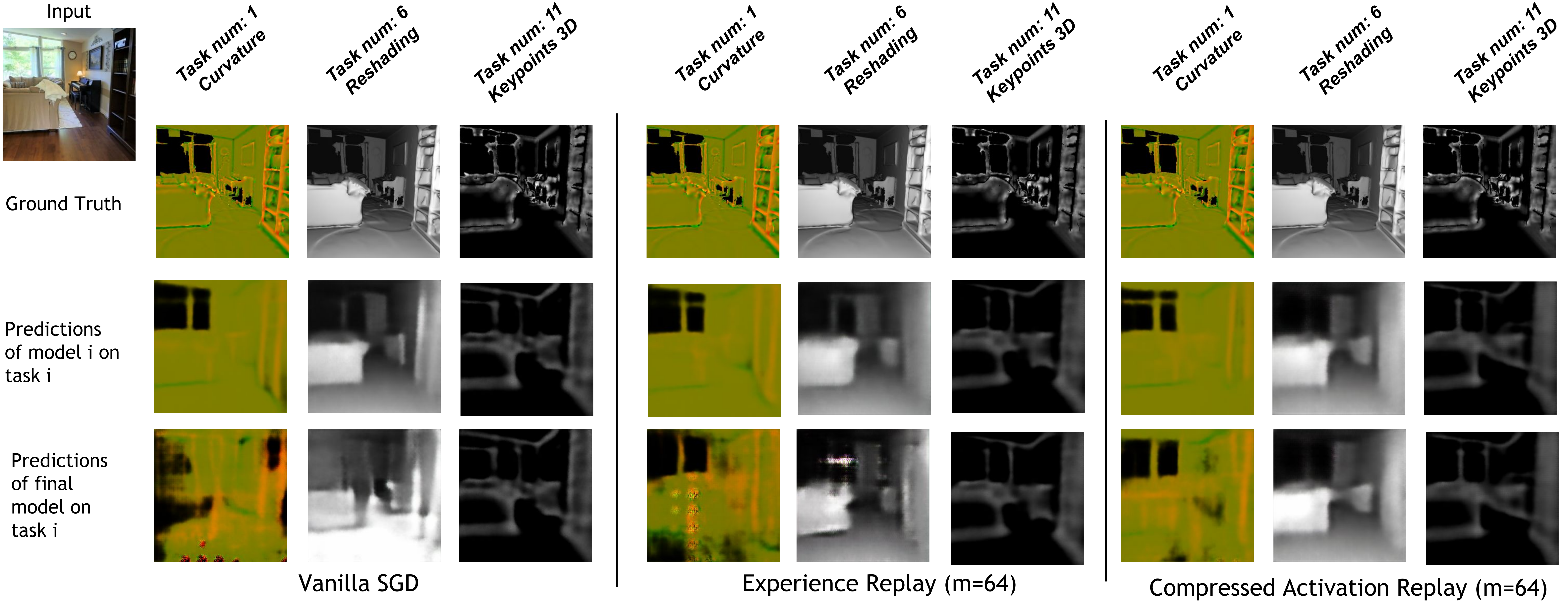}
\caption{\small Visualizations of predictions made on Taskonomy dataset. Compressed Activation Replay (CAR) drives the predictions to higher quality, compared with vanilla SGD and ER.}
\vspace{-4mm}
\label{fig:prediction_vis} 
\end{figure*}
\subsection{Taskonomy}
\vspace{-2mm}
We compare the proposed CAR approach with multiple baseline methods in the Taskonomy dataset mentioned in Section~\ref{sec:vanilla_er}: vanilla SGD, A-GEM, and vanilla experience replay (ER).

\begin{table*}
\caption{\small\textbf{Taskonomy results} using various replay methods. Forgetting (\textit{fgt}$\pm$std) and performance drop ($\Delta$\textit{perf}$\pm$std) values are reported (smaller values are better). All results are averaged over $3$ random seeds.}
\vskip -1em
\label{tab:taskonomy}
\setlength\tabcolsep{3pt}
\begin{center}
\resizebox{\textwidth}{!}{
\begin{tabular}{r|cc|cc|cc|cc}
\toprule
Method & \multicolumn{2}{c|}{$m=64$}  & \multicolumn{2}{c|}{$m=128$} & \multicolumn{2}{c|}{$m=256$} & \multicolumn{2}{c}{$m=512$} \\  
 & \textit{fgt} ($\downarrow$) & $\Delta$\textit{perf} ($\downarrow$) & \textit{fgt} ($\downarrow$) & $\Delta$\textit{perf} ($\downarrow$) & \textit{fgt} ($\downarrow$) & $\Delta$\textit{perf} ($\downarrow$) & \textit{fgt} ($\downarrow$) & $\Delta$\textit{perf} ($\downarrow$) \\
 \midrule 
 \multicolumn{9}{l}{\it {Sequence}: \small \textit{Curv, Cls (obj), Normal, Seg, Depth, Reshade, Edge (tex), Cls (sc), Keypts2d, Edge (occ), Keypts3d}} \\
\midrule
 SGD  & 191.6\std{13.8} & 101.6\std{9.2} & 191.6\std{13.8} & 101.6\std{9.2} & 191.6\std{13.8} & 101.6\std{9.2} & 191.6\std{13.8} & 101.6\std{9.2} \\
 A-GEM   &  43.6\std{0.7} & 48.0\std{0.9} & 44.1\std{2.3} & 50.2\std{2.8} & 38.9\std{1.9} & 49.0\std{2.9} & 24.9\std{3.1} & 50.9\std{4.7} \\
 ER     &  40.2\std{1.9} & 20.7\std{0.5} & 24.7\std{1.4} & 13.8\std{0.7} & 14.7\std{3.1} & 8.5\std{0.5} & 8.3\std{0.8} &  5.4\std{2.4} \\
CAR (ours) &   13.4\std{3.6} & 23.3\std{2.8} & 6.3\std{0.7} & 19.3\std{3.1} & 4.3\std{0.2} & 12.4\std{3.8} &  2.2\std{0.8} & 10.8\std{2.2} \\
 \midrule
 \multicolumn{9}{l}{\it {Sequence}: \small\textit{Depth, Keypts2d, Seg, Normal, Cls (obj), Curv, Edge (occ), Reshade, Cls (sc), Keypts3d, Edge (tex)}} \\
\midrule
 SGD  & 110.5\std{16.3} & 79.5\std{14.3} & 110.5\std{16.3} & 79.5\std{14.3} & 110.5\std{16.3} & 79.5\std{14.3} & 110.5\std{16.3} & 79.5\std{14.3} \\
 A-GEM   &  46.5\std{4.3} & 64.0\std{4.9} & 48.9\std{3.0} & 62.3\std{4.0} & 46.7\std{4.8} & 71.7\std{4.9} & 32.6\std{5.1} & 60.1\std{5.7} \\
 ER  &  43.8\std{2.7} & 33.4\std{3.2} & 24.0\std{4.5} & 24.3\std{3.2} & 19.9\std{6.4} & 12.7\std{4.5} & 9.1\std{3.9} &  5.3\std{2.9} \\
CAR (ours) & 15.9\std{4.6} & 35.6\std{3.1} & 7.9\std{2.6} & 26.8\std{4.5} & 6.7\std{1.2} & 14.3\std{0.7} & 5.7\std{3.1} & 10.8\std{1.1} \\
\bottomrule
\end{tabular}
}
\vspace{-4mm}
\end{center}
\end{table*}

Two diferent task sequences are used in the evaluation. Results are shown in Table \ref{tab:taskonomy}. The trend in both task sequences are consistent where CAR significantly outperforms the other methods on the forgetting measure, \textit{e.g.}, it reduces forgetting rate of vanilla ER by $28\%$ when $m=64$. The performance drop of CAR is slightly higher but still comparable with ER. This can be explained by the stability-plasticity dilemma in continual learning~\citep{mermillod2013stability,mirzadeh2020understanding}, where, stability against forgetting is achieved at the slight cost in the plasticity of the model, \textit{i.e.}, the ability of  quickly acquiring new knowledge. The proposed feature matching loss acts as a regularizer and stabilizes the model resulting in a significant decrease in forgetting traded by a relatively smaller drop in learning new tasks. Nevertheless, the parameter $\lambda_{fm}$ is designed to control this trade-off and tune it according to the application. According to the tables, bigger memory size also improves the performance for the proposed CAR method. We visualize the model predictions  in Figure~\ref{fig:prediction_vis}, which show that the images predicted by CAR have higher quality than vanilla SGD and ER.

\vspace{-2mm}
\subsection{Split-CIFAR and Split-miniImageNet}
\vspace{-2mm}
In addition to the large-scale Taskonomy dataset, we perform experiments on two continual learning datasets - Split-CIFAR and Split-miniImageNet. We follow the protocol in \cite{chaudhry2020hindsightreplay, chaudhry2018efficient} for all the experiments -- Of the $20$ tasks in the task sequence, $3$ tasks are used for validation, while the other $17$ tasks are used to report the final performance. Each task is trained for a single epoch with a batch size of $10$.
Since each task is a $5$-way classification problem, we train models using cross-entropy loss. We use the reduced Resnet-18 model as the encoder following \cite{chaudhry2020hindsightreplay, chaudhry2018efficient}. Task heads are one-layer linear networks mapping the encoder layer to the logits.

In addition to the baselines listed previously, we also compare with the three additional replay based methods whose performance on Split-CIFAR and Split-miniImageNet were reported in the literature, i.e., iCaRL~\citep{Rebuffi_2017}, MER~\citep{riemer2018learning}, HAL~\citep{chaudhry2020hindsightreplay}.

\begin{table*}
\caption{\small\textbf{Split-CIFAR and Split-miniImageNet results:}  Average accuracy (acc$\pm$std \%) and forgetting ($\pm$std) values are reported. Results are averaged over $5$ random seeds. The best result is highlighted in bold.}\vskip -1em
\label{tab:split_cifar}
\begin{center}
\resizebox{\textwidth}{!}{
\begin{tabular}{r|cc|cc|cc|cc}
\toprule
& \multicolumn{4}{|c|}{\sc Split-CIFAR} & \multicolumn{4}{c}{\sc Split-miniImageNet}\\
Method & \multicolumn{2}{c}{$m=85$}  & \multicolumn{2}{c|}{$m=425$} &\multicolumn{2}{c}{$m=85$}  &  \multicolumn{2}{c}{$m=425$} \\  
 & \it {acc}($\uparrow$) & \it fgt($\downarrow$) & \it acc($\uparrow$) & \it fgt($\downarrow$) & \it acc($\uparrow$) & \it fgt($\downarrow$) & \it acc($\uparrow$)  & \it fgt($\downarrow
 $) \\
 \midrule \midrule 
 EWC    & 42.4\std{3.0} & 0.25\std{0.02} & - & -  & 37.7\std{3.3} & 0.21\std{0.03} & - & -\\

 iCaRL  & 46.4\std{1.2} & 0.16\std{0.01} & 51.2\std{1.3} & 0.13\std{0.02} & - & - & - & - \\
 A-GEM   & 54.9\std{2.9} & 0.14\std{0.03} & 59.9\std{2.6} & 0.10\std{0.02} & 48.2\std{2.4} & 0.13\std{0.02}  & 54.3\std{1.6} & 0.08\std{0.01}\\
 MER    & 49.7\std{2.9} & 0.19\std{0.03} & 60.6\std{2.1} & 0.09\std{0.02} & 45.4\std{1.4} & 0.15\std{0.01}  & 54.8\std{1.7} & 0.07\std{0.01} \\
 ER     & 56.2\std{1.9} & 0.13\std{0.01} & 62.6\std{1.8} & 0.06\std{0.01} & 49.0\std{2.6} & 0.12\std{0.02}  & 54.2\std{3.2} & 0.08\std{0.02} \\
 HAL    & 60.4\std{0.5} & 0.10\std{0.01} & 64.4\std{2.1} & 0.06\std{0.01} & 51.6\std{2.0} & 0.10\std{0.01} & \bf 57.2\std{1.5} & \bf 0.06\std{0.01}\\
 \midrule
 CAR (ours)  & \bf 62.4\std{1.5} & \bf 0.06\std{0.01} & \bf 66.5\std{1.8} &  \bf 0.04\std{0.01} & \bf 54.2\std{2.9} & \bf 0.06\std{0.01} & 56.2\std{1.8} & \bf 0.06\std{0.01} \\
\bottomrule
\end{tabular}
}
\vspace{-4mm}
\end{center}
\end{table*}

The results on Split-CIFAR and Split-miniImageNet are shown in Table \ref{tab:split_cifar}. We found Compressed Activation Replay achieves significant improvement in CIFAR in terms of both final accuracy and forgetting rate. For Split-miniImageNet, our method outperforms all the methods significantly in the small memory case, and performs comparably to HAL in large memory case.

\subsection{Discussion}\label{sec:ablation}

\textbf{Compression techniques.} First, we compare the performance of different compression techniques used in CAR framework. Let the dimension of feature tensor be $(n_{f} \times w \times h)$. We employ the following simple compression schemes: (a) \textit{Spatial}: average pooling along the spatial dimension (dimension $n_{f}$); (b) \textit{Channel}: average pooling along the channel dimension (dimension $wh$); (c) \textit{Spatial + Channel}: a concatenation of spatial and channel pooling (dimension $n_f+wh$).

\begin{table*}
\caption{\small \textbf{Taskonomy compression techniques:} Forgetting (\textit{fgt}), performance drop ($\Delta$\textit{perf}) and memory overhead (\textit{mem}) values are reported. All results are averaged over $3$ random seeds.}
\vskip -1em
\label{tab:taskonomy_compression}
\begin{center}
\resizebox{\textwidth}{!}{
\begin{tabular}{r|ccc|ccc|ccc|ccc}
\toprule
Compression & \multicolumn{3}{|c}{$m=64$}  & \multicolumn{3}{|c}{$m=128$} & \multicolumn{3}{|c}{$m=256$} & \multicolumn{3}{|c}{$m=512$} \\  
 Method& \textit{fgt} & $\Delta$\textit{perf} & \textit{mem} & \textit{fgt} & $\Delta$\textit{perf}  & \textit{mem} & \textit{fgt} & $\Delta$\textit{perf} & \textit{mem} & \textit{fgt}  & $\Delta$\textit{perf} &  \textit{mem} \\
 \midrule 
 \midrule
 No compression  & 9.18 & 29.45 & 335.5M & 4.33 & 22.73 & 671.0 M & 0.77 & 22.06 & 1.3 G & 0.51  & 15.44 & 2.7 G \\
 \midrule
 Spatial & 18.15 & 31.45 & 5.1 M & 9.78 & 23.58 & 10.2 M & 7.75 & 16.31 & 20.4 M & 4.30 & 12.53 & 41.0 M \\
 Channel & 18.00 & 18.25 & 0.2 M & 21.38 & 33.22 & 0.3 M & 6.83  & 28.40 & 0.6 M & 6.47 & 28.28 & 1.3 M \\
 Spatial + Channel   & 12.43 & 25.97 & 5.4 M & 6.25  & 19.32 & 10.8 M & 4.29 & 15.62 & 21.6 M & 2.18 & 10.80 & 43.3 M \\
\bottomrule
\end{tabular}
}
\vspace{-4mm}
\end{center}
\end{table*}

In addition to performance, we report in Table.~\ref{tab:taskonomy_compression} the memory overhead needed for storing the samples in the replay buffer. With significantly lower memory overhead, compression techniques perform comparably to those without compression. Among the compression methods, channel pooling performs poorly compared to spatial pooling. Spatial+Channel pooling achieves the best result.

\textbf{Regularization functions.}
We compare different regularization functions $\ell_{fm}$ for feature matching. Natural choices are $\ell_2$ and $\ell_1$ distances. Besides that we also experiment with weighted version of $\ell_1$ and $\ell_2$ where the weights are computed by accumulating the gradients, since neurons with high magnitude of changes are often important to the corresponding tasks. In addition, we also tried the MMD loss~\citep{gretton2007kernel}, a popular distribution matching distance. Interestingly, none of them is significantly better than the rest. Detailed results can be found in Appendix.

\vspace{-2mm}
\section{Conclusion}
\vspace{-2mm}
We studied the large scale continual learning using memory replay based methods on a non-conventional diverse set of tasks.
We compared the performance of different  methods in both Taskonomy and medium-scale CIFAR/miniImageNet datasets and found that vanilla experience replay is a strong baseline algorithm despite its simplicity. However, feature drift is observed while training with a standard replay memory.
We further proposed the Compressed Activation Replay framework, which aims to reduce such drifting. Experimental results show that our approach achieves the state-of-the-art performance on both large scale Taskonomy benchmark and standard medium-size continual learning datasets.
\section*{Acknowledgement}
The authors would like to thank Dilan Gorur, Polina Kirichenko, Razvan Pascanu for their helpful discussions and meaningful feedbacks.

\bibliography{iclr2021_conference}

\begin{thebibliography}{39}
\providecommand{\natexlab}[1]{#1}
\providecommand{\url}[1]{\texttt{#1}}
\expandafter\ifx\csname urlstyle\endcsname\relax
  \providecommand{\doi}[1]{doi: #1}\else
  \providecommand{\doi}{doi: \begingroup \urlstyle{rm}\Url}\fi

\bibitem[Adel et~al.(2020)Adel, Zhao, and Turner]{Adel2020ContinualLW}
Tameem Adel, Han Zhao, and R.~Turner.
\newblock Continual learning with adaptive weights (claw).
\newblock \emph{ArXiv}, abs/1911.09514, 2020.

\bibitem[Aljundi et~al.(2019)Aljundi, Lin, Goujaud, and
  Bengio]{aljundi2019gradient}
Rahaf Aljundi, Min Lin, Baptiste Goujaud, and Yoshua Bengio.
\newblock Gradient based sample selection for online continual learning.
\newblock In \emph{Advances in Neural Information Processing Systems}, pp.\
  11816--11825, 2019.

\bibitem[Borsos et~al.(2020)Borsos, Mutn{\`y}, and Krause]{borsos2020coresets}
Zal{\'a}n Borsos, Mojm{\'\i}r Mutn{\`y}, and Andreas Krause.
\newblock Coresets via bilevel optimization for continual learning and
  streaming.
\newblock \emph{arXiv preprint arXiv:2006.03875}, 2020.

\bibitem[Chaudhry et~al.(2018)Chaudhry, Ranzato, Rohrbach, and
  Elhoseiny]{chaudhry2018efficient}
Arslan Chaudhry, Marc'Aurelio Ranzato, Marcus Rohrbach, and Mohamed Elhoseiny.
\newblock Efficient lifelong learning with {A-GEM}.
\newblock \emph{arXiv preprint arXiv:1812.00420}, 2018.

\bibitem[Chaudhry et~al.(2019)Chaudhry, Rohrbach, Elhoseiny, Ajanthan, Dokania,
  Torr, and Ranzato]{chaudhry2019tiny}
Arslan Chaudhry, Marcus Rohrbach, Mohamed Elhoseiny, Thalaiyasingam Ajanthan,
  Puneet~K Dokania, Philip~HS Torr, and Marc'Aurelio Ranzato.
\newblock On tiny episodic memories in continual learning.
\newblock \emph{arXiv preprint arXiv:1902.10486}, 2019.

\bibitem[Chaudhry et~al.(2020)Chaudhry, Gordo, Dokania, Torr, and
  Lopez-Paz]{chaudhry2020hindsightreplay}
Arslan Chaudhry, Albert Gordo, Puneet~K Dokania, Philip Torr, and David
  Lopez-Paz.
\newblock Using hindsight to anchor past knowledge in continual learning.
\newblock \emph{arXiv preprint arXiv:2002.08165}, 2020.

\bibitem[Chrysakis \& Moens(2020)Chrysakis and Moens]{chrysakis2020online}
Aristotelis Chrysakis and Marie-Francine Moens.
\newblock Online continual learning from imbalanced data.
\newblock \emph{Proceedings of Machine Learning and Systems}, pp.\  8303--8312,
  2020.

\bibitem[Farajtabar et~al.(2020)Farajtabar, Azizan, Mott, and
  Li]{farajtabar2019orthogonal}
Mehrdad Farajtabar, Navid Azizan, Alex Mott, and Ang Li.
\newblock Orthogonal gradient descent for continual learning.
\newblock In \emph{International Conference on Artificial Intelligence and
  Statistics}, 2020.

\bibitem[Gretton et~al.(2007)Gretton, Borgwardt, Rasch, Sch{\"o}lkopf, and
  Smola]{gretton2007kernel}
Arthur Gretton, Karsten Borgwardt, Malte Rasch, Bernhard Sch{\"o}lkopf, and
  Alex~J Smola.
\newblock A kernel method for the two-sample-problem.
\newblock In \emph{Advances in neural information processing systems}, pp.\
  513--520, 2007.

\bibitem[Hayes et~al.(2019)Hayes, Cahill, and Kanan]{Hayes_2019}
Tyler~L. Hayes, Nathan~D. Cahill, and Christopher Kanan.
\newblock Memory efficient experience replay for streaming learning.
\newblock \emph{2019 International Conference on Robotics and Automation
  (ICRA)}, May 2019.
\newblock \doi{10.1109/icra.2019.8793982}.
\newblock URL \url{http://dx.doi.org/10.1109/ICRA.2019.8793982}.

\bibitem[Isele \& Cosgun(2018)Isele and Cosgun]{isele2018selective}
David Isele and Akansel Cosgun.
\newblock Selective experience replay for lifelong learning.
\newblock \emph{arXiv preprint arXiv:1802.10269}, 2018.

\bibitem[Kirkpatrick et~al.(2017)Kirkpatrick, Pascanu, Rabinowitz, Veness,
  Desjardins, Rusu, Milan, Quan, Ramalho, Grabska-Barwinska, and
  et~al.]{Kirkpatrick_2017}
James Kirkpatrick, Razvan Pascanu, Neil Rabinowitz, Joel Veness, Guillaume
  Desjardins, Andrei~A. Rusu, Kieran Milan, John Quan, Tiago Ramalho, Agnieszka
  Grabska-Barwinska, and et~al.
\newblock Overcoming catastrophic forgetting in neural networks.
\newblock \emph{Proceedings of the National Academy of Sciences}, 114\penalty0
  (13):\penalty0 3521–3526, Mar 2017.
\newblock ISSN 1091-6490.
\newblock \doi{10.1073/pnas.1611835114}.
\newblock URL \url{http://dx.doi.org/10.1073/pnas.1611835114}.

\bibitem[Knoblauch et~al.(2020)Knoblauch, Husain, and
  Diethe]{knoblauch2020optimal}
Jeremias Knoblauch, Hisham Husain, and Tom Diethe.
\newblock Optimal continual learning has perfect memory and is {NP}-hard.
\newblock \emph{arXiv preprint arXiv:2006.05188}, 2020.

\bibitem[Li et~al.(2019)Li, Zhou, Wu, Socher, and Xiong]{li2019learn}
Xilai Li, Yingbo Zhou, Tianfu Wu, Richard Socher, and Caiming Xiong.
\newblock Learn to grow: A continual structure learning framework for
  overcoming catastrophic forgetting.
\newblock \emph{arXiv preprint arXiv:1904.00310}, 2019.

\bibitem[Li \& Hoiem(2017)Li and Hoiem]{li2017learning}
Zhizhong Li and Derek Hoiem.
\newblock Learning without forgetting.
\newblock \emph{IEEE transactions on pattern analysis and machine
  intelligence}, 40\penalty0 (12):\penalty0 2935--2947, 2017.

\bibitem[Lomonaco \& Maltoni(2017)Lomonaco and Maltoni]{lomonaco2017core50}
Vincenzo Lomonaco and Davide Maltoni.
\newblock {CORe50}: a new dataset and benchmark for continuous object
  recognition.
\newblock \emph{arXiv preprint arXiv:1705.03550}, 2017.

\bibitem[Lopez-Paz \& Ranzato(2017)Lopez-Paz and Ranzato]{lopez2017gradient}
David Lopez-Paz and Marc'Aurelio Ranzato.
\newblock Gradient episodic memory for continual learning.
\newblock In \emph{Advances in neural information processing systems}, pp.\
  6467--6476, 2017.

\bibitem[Mermillod et~al.(2013)Mermillod, Bugaiska, and
  Bonin]{mermillod2013stability}
Martial Mermillod, Aur{\'e}lia Bugaiska, and Patrick Bonin.
\newblock The stability-plasticity dilemma: Investigating the continuum from
  catastrophic forgetting to age-limited learning effects.
\newblock \emph{Frontiers in psychology}, 4:\penalty0 504, 2013.

\bibitem[Mirzadeh et~al.(2020)Mirzadeh, Farajtabar, and
  Ghasemzadeh]{mirzadeh2020dropout}
Seyed-Iman Mirzadeh, Mehrdad Farajtabar, and Hassan Ghasemzadeh.
\newblock Dropout as an implicit gating mechanism for continual learning.
\newblock In \emph{Proceedings of the IEEE/CVF Conference on Computer Vision
  and Pattern Recognition Workshops}, pp.\  232--233, 2020.

\bibitem[{Mirzadeh} et~al.(2020){Mirzadeh}, {Farajtabar}, {Pascanu}, and
  {Ghasemzadeh}]{mirzadeh2020understanding}
Seyed~Iman {Mirzadeh}, Mehrdad {Farajtabar}, Razvan {Pascanu}, and Hassan
  {Ghasemzadeh}.
\newblock Understanding the role of training regimes in continual learning.
\newblock \emph{arXiv preprint arXiv:2006.06958}, 2020.

\bibitem[Mnih et~al.(2013)Mnih, Kavukcuoglu, Silver, Graves, Antonoglou,
  Wierstra, and Riedmiller]{mnih2013playing}
Volodymyr Mnih, Koray Kavukcuoglu, David Silver, Alex Graves, Ioannis
  Antonoglou, Daan Wierstra, and Martin Riedmiller.
\newblock Playing atari with deep reinforcement learning.
\newblock \emph{arXiv preprint arXiv:1312.5602}, 2013.

\bibitem[Nguyen et~al.(2017)Nguyen, Li, Bui, and Turner]{nguyen2017variational}
Cuong~V Nguyen, Yingzhen Li, Thang~D Bui, and Richard~E Turner.
\newblock Variational continual learning.
\newblock \emph{arXiv preprint arXiv:1710.10628}, 2017.

\bibitem[Pomponi et~al.(2020)Pomponi, Scardapane, Lomonaco, and
  Uncini]{pomponi2020efficient}
Jary Pomponi, Simone Scardapane, Vincenzo Lomonaco, and Aurelio Uncini.
\newblock Efficient continual learning in neural networks with embedding
  regularization.
\newblock \emph{Neurocomputing}, 2020.

\bibitem[Rebuffi et~al.(2017)Rebuffi, Kolesnikov, Sperl, and
  Lampert]{Rebuffi_2017}
Sylvestre-Alvise Rebuffi, Alexander Kolesnikov, Georg Sperl, and Christoph~H.
  Lampert.
\newblock {iCaRL}: Incremental classifier and representation learning.
\newblock \emph{2017 IEEE Conference on Computer Vision and Pattern Recognition
  (CVPR)}, Jul 2017.
\newblock \doi{10.1109/cvpr.2017.587}.
\newblock URL \url{http://dx.doi.org/10.1109/CVPR.2017.587}.

\bibitem[Riemer et~al.(2018)Riemer, Cases, Ajemian, Liu, Rish, Tu, and
  Tesauro]{riemer2018learning}
Matthew Riemer, Ignacio Cases, Robert Ajemian, Miao Liu, Irina Rish, Yuhai Tu,
  and Gerald Tesauro.
\newblock Learning to learn without forgetting by maximizing transfer and
  minimizing interference.
\newblock \emph{arXiv preprint arXiv:1810.11910}, 2018.

\bibitem[Ritter et~al.(2018)Ritter, Botev, and Barber]{ritter2018online}
Hippolyt Ritter, Aleksandar Botev, and David Barber.
\newblock Online structured {L}aplace approximations for overcoming
  catastrophic forgetting.
\newblock In \emph{Advances in Neural Information Processing Systems}, pp.\
  3738--3748, 2018.

\bibitem[Robins(1995)]{robins1995catastrophic}
Anthony Robins.
\newblock Catastrophic forgetting, rehearsal and pseudorehearsal.
\newblock \emph{Connection Science}, 7\penalty0 (2):\penalty0 123--146, 1995.

\bibitem[Rolnick et~al.(2019)Rolnick, Ahuja, Schwarz, Lillicrap, and
  Wayne]{rolnick2019experience}
David Rolnick, Arun Ahuja, Jonathan Schwarz, Timothy Lillicrap, and Gregory
  Wayne.
\newblock Experience replay for continual learning.
\newblock In \emph{Advances in Neural Information Processing Systems}, pp.\
  350--360, 2019.

\bibitem[Rusu et~al.(2016)Rusu, Rabinowitz, Desjardins, Soyer, Kirkpatrick,
  Kavukcuoglu, Pascanu, and Hadsell]{rusu2016progressive}
Andrei~A Rusu, Neil~C Rabinowitz, Guillaume Desjardins, Hubert Soyer, James
  Kirkpatrick, Koray Kavukcuoglu, Razvan Pascanu, and Raia Hadsell.
\newblock Progressive neural networks.
\newblock \emph{arXiv preprint arXiv:1606.04671}, 2016.

\bibitem[Schaul et~al.(2015)Schaul, Quan, Antonoglou, and
  Silver]{schaul2015prioritized}
Tom Schaul, John Quan, Ioannis Antonoglou, and David Silver.
\newblock Prioritized experience replay.
\newblock \emph{arXiv preprint arXiv:1511.05952}, 2015.

\bibitem[Shin et~al.(2017)Shin, Lee, Kim, and Kim]{shin2017continual}
Hanul Shin, Jung~Kwon Lee, Jaehong Kim, and Jiwon Kim.
\newblock Continual learning with deep generative replay.
\newblock In \emph{Advances in Neural Information Processing Systems}, pp.\
  2990--2999, 2017.

\bibitem[van~de Ven \& Tolias(2019)van~de Ven and Tolias]{van2019three}
Gido~M van~de Ven and Andreas~S Tolias.
\newblock Three scenarios for continual learning.
\newblock \emph{arXiv preprint arXiv:1904.07734}, 2019.

\bibitem[van~de Ven et~al.(2020)van~de Ven, Siegelmann, and
  Tolias]{van2020brain}
Gido~M van~de Ven, Hava~T Siegelmann, and Andreas~S Tolias.
\newblock Brain-inspired replay for continual learning with artificial neural
  networks.
\newblock \emph{Nature communications}, 11\penalty0 (1):\penalty0 1--14, 2020.

\bibitem[van~der Maaten \& Hinton(2008)van~der Maaten and Hinton]{tsne}
Laurens van~der Maaten and Geoffrey Hinton.
\newblock Visualizing data using {t-SNE}.
\newblock \emph{Journal of Machine Learning Research}, 9:\penalty0 2579--2605,
  2008.
\newblock URL \url{http://www.jmlr.org/papers/v9/vandermaaten08a.html}.

\bibitem[Yin et~al.(2020)Yin, Farajtabar, and Li]{yin2020sola}
Dong Yin, Mehrdad Farajtabar, and Ang Li.
\newblock {SOLA}: Continual learning with second-order loss approximation.
\newblock \emph{arXiv preprint arXiv:2006.10974}, 2020.

\bibitem[Yoon et~al.(2018)Yoon, Yang, Lee, and Hwang]{yoon2018lifelong}
Jaehong Yoon, Eunho Yang, Jungtae Lee, and Sung~Ju Hwang.
\newblock Lifelong learning with dynamically expandable networks.
\newblock In \emph{International Conference on Learning Representations}. ICLR,
  2018.

\bibitem[Zamir et~al.(2018)Zamir, Sax, Shen, Guibas, Malik, and
  Savarese]{zamir2018taskonomy}
Amir~R Zamir, Alexander Sax, William Shen, Leonidas~J Guibas, Jitendra Malik,
  and Silvio Savarese.
\newblock Taskonomy: Disentangling task transfer learning.
\newblock In \emph{Proceedings of the IEEE conference on computer vision and
  pattern recognition}, pp.\  3712--3722, 2018.

\bibitem[Zenke et~al.(2017)Zenke, Poole, and Ganguli]{zenke2017continual}
Friedemann Zenke, Ben Poole, and Surya Ganguli.
\newblock Continual learning through synaptic intelligence.
\newblock In \emph{Proceedings of the 34th International Conference on Machine
  Learning-Volume 70}, pp.\  3987--3995. JMLR, 2017.

\bibitem[Zhang et~al.(2020)Zhang, Sax, Zamir, Guibas, and
  Malik]{sidetuning2020}
Jeffrey~O. Zhang, Alexander Sax, Amir Zamir, Leonidas~J. Guibas, and Jitendra
  Malik.
\newblock Side-{T}uning: Network adaptation via additive side networks.
\newblock In \emph{European Conference on Computer Vision (ECCV)}, 2020.

\end{thebibliography}
\bibliographystyle{iclr2021_conference}

\appendix
\newpage

\section{Regularization functions}
In this experiment, we perform an ablation study to understand the effect of regularization functions $\ell_{fm}(\cdot)$. We use the following loss functions.
\begin{enumerate}
    \item $\ell_{1}$: Measures the $\ell_{1}$ distance between the feature maps.
    \item $\ell_{2}$: Measures the $\ell_{2}$ distance between the feature maps.
    \item $\ell_{1} + \ell_{2}$: Use a combination of $\ell_1$ and $\ell_2$ losses.
    \item Weighted $\ell_{1}$: We compute a weighted $\ell_1$ loss, where weights are computed using averaged gradients of loss with respect to embeddings i.e., $\nabla\ell_{\textbf{f}}$. Larger the gradient magnitudes, more influential are the feature components. We would like to point out that this loss is similar to the Fisher information used in EWC~\cite{Kirkpatrick_2017}.
    \item Weighted $\ell_{2}$: Weighted version of $\ell_2$ loss.
    \item MMD: Use of MMD loss for measuring the distributional distance between the input feature maps.
\end{enumerate}

The results of using these different regularization functions on Split-CIFAR and Split-miniImageNet datasets are shown in Table.~\ref{tab:split_cls_distribution_matching}. We observe that all techniques except MMD perform comparable to each other. So, for the ease of implementation, we use $\ell_2$ in all our experiments. MMD results in low performance. 
\begin{table*}[ht]
\caption{\textbf{Distribution Matching Techniques}  Average accuracy (in \%) and forgetting values are reported. All results are averaged over $5$ random seeds.}
\label{tab:split_cls_distribution_matching}
\begin{center}
\resizebox{0.7\textwidth}{!}{
\begin{tabular}{l|cc|cc|cc|cc}
\toprule
 & \multicolumn{4}{|c|}{Split-CIFAR} & \multicolumn{4}{|c}{Split-miniImageNet} \\  
Method & \multicolumn{2}{|c|}{$m=85$}  & \multicolumn{2}{|c|}{$m=425$} & \multicolumn{2}{|c}{$m=85$} & \multicolumn{2}{|c}{$m=425$} \\  
 & \textit{acc} & \textit{fgt} & \textit{acc} & \textit{fgt} & \textit{acc} & \textit{fgt} & \textit{acc} & \textit{fgt} \\
\midrule
 $\ell_{1}$ & 62.4 & 0.07 & 66.1 & 0.04 & 54.1 & 0.07 & 55.4 & 0.07   \\
 $\ell_{2}$ & 62.4 & 0.07 & 66.4 & 0.04 & 53.8 & 0.06 & 56.3 & 0.06 \\
 $\ell_{1} + \ell_{2}$ & 62.0 & 0.07 & 66.1 & 0.04 & 53.5 & 0.07 & 56.4 & 0.06  \\
 Weighted $\ell_{1}$ &  63.2 & 0.06 & 67.1 &  0.04 & 54.6 & 0.06 & 55.8 & 0.07   \\
 Weighted $\ell_{2}$ & 61.5 & 0.07 & 67.0 & 0.04 & 53.8 & 0.06 & 56.3 & 0.07   \\
 MMD & 56.4 & 0.11 & 63.2 & 0.06 & 51.2 & 0.08 & 55.0 & 0.08 \\
\bottomrule
\end{tabular}
}
\end{center}
\end{table*}

\section{Intermediate layer activation matching}
The compressed activation matching framework performs activation mapping of feature representations. In all experiments reported in the paper, we used the output of encoder for this task. What happens if we use the representations of the previous layers? To study this, we perform compressed activation matching on intermediate representations of the encoder in addition to the final embedding layer. More concretely, we take the outputs of each of the $4$ Resnet blocks of Resnet-50 network, perform feature compression and concatenate to obtain a single representation. We then use this concatenated representation to perform feature matching. The results of intermediate feature matching are shown in Table.~\ref{tab:taskonomy_intermediate}. We observe that we do not obtain any gains from the intermediate feature matching.

\begin{table*}[ht]
\caption{\textbf{Effect of using intermediate representations:} Forgetting (\textit{fgt}) and performance drop ($\Delta$\textit{perf}) values are reported. All results are averaged over $3$ random seeds.}
\label{tab:taskonomy_intermediate}
\begin{center}
\resizebox{\textwidth}{!}{
\begin{tabular}{l |cc|cc|cc|cc}
\toprule
Method & \multicolumn{2}{|c|}{$m=64$}  & \multicolumn{2}{|c|}{$m=128$} & \multicolumn{2}{|c}{$m=256$} & \multicolumn{2}{|c}{$m=512$} \\  
 & \textit{fgt} ($\downarrow$) & $\Delta$\textit{perf} ($\downarrow$) & \textit{fgt} ($\downarrow$) & $\Delta$\textit{perf} ($\downarrow$) & \textit{fgt} ($\downarrow$) & $\Delta$\textit{perf} ($\downarrow$) & \textit{fgt} ($\downarrow$) & $\Delta$\textit{perf} ($\downarrow$) \\
 \midrule 
 \midrule
 CAR  & 13.37 & 23.25 & 4.80 & 17.32 & 2.98 & 12.43 & 1.91 & 11.51 \\
 CAR (\textit{intermediate}) & 15.44 & 22.17 & 8.75 & 19.01 & 3.53 & 14.84 & 0.58 & 13.59 \\
\bottomrule
\end{tabular}
}
\end{center}
\end{table*}

\section{Sensitivity of $\lambda_{fm}$}
\begin{table*}[ht]
\caption{\textbf{Sensitivity of $\lambda_{fm}$:} Forgetting (\textit{fgt}) and performance drop ($\Delta$\textit{perf}) values are reported. All results are averaged over $3$ random seeds.}
\label{tab:taskonomy_sensitivity_analysis}
\begin{center}
\resizebox{\textwidth}{!}{
\begin{tabular}{l |cc|cc|cc|cc}
\toprule
Method & \multicolumn{2}{|c|}{$m=64$}  & \multicolumn{2}{|c|}{$m=128$} & \multicolumn{2}{|c}{$m=256$} & \multicolumn{2}{|c}{$m=512$} \\  
 & \textit{fgt} ($\downarrow$) & $\Delta$\textit{perf} ($\downarrow$) & $\Delta$\textit{fgt} ($\downarrow$) & $\Delta$\textit{perf} ($\downarrow$) & \textit{fgt} ($\downarrow$) & $\Delta$\textit{perf} ($\downarrow$) & \textit{fgt} ($\downarrow$) & $\Delta$\textit{perf} ($\downarrow$) \\
 \midrule 
 \midrule
 $\lambda_{fm} = 0.25$ & 32.57 & 25.36 & 24.09 & 19.06 & 15.79 & 11.49 & 9.06 & 13.84\\
 $\lambda_{fm} = 1.0$  & 28.49 & 25.33 & 20.45 & 20.91 & 11.24 & 17.83 & 7.09 & 12.89 \\
 $\lambda_{fm} = 5.0$  & 25.19 & 29.06 & 12.01 & 18.46 & 8.26 & 17.03 &  2.18 & 10.44 \\
 $\lambda_{fm} = 20.0$ & 18.21 & 24.72 & 10.09 & 18.86 & 4.29 & 12.43 &  -0.29 & 14.94 \\
 $\lambda_{fm} = 50.0$ & 13.37 & 23.25 & 6.25 & 19.32  & 2.13 & 20.17 &  -1.65 & 17.29\\
 
\bottomrule
\end{tabular}
}
\end{center}
\end{table*}


\begin{table*}[ht]
\caption{\textbf{Sensitivity of $\lambda_{fm}$:} Average accuracy (in \%) and forgetting values are reported. All results are averaged over $5$ random seeds.}
\label{tab:split_cifar_sensitivity}
\begin{center}
\resizebox{0.7\textwidth}{!}{
\begin{tabular}{l|cc|cc|cc|cc}
\toprule
 & \multicolumn{4}{|c|}{Split-CIFAR} & \multicolumn{4}{|c}{Split-miniImagenet} \\  
Method & \multicolumn{2}{|c|}{$m=85$}  & \multicolumn{2}{|c|}{$m=425$} & \multicolumn{2}{|c}{$m=85$} & \multicolumn{2}{|c}{$m=425$} \\  
 & \textit{acc} & \textit{fgt} & \textit{acc} & \textit{fgt} & \textit{acc} & \textit{fgt} & \textit{acc} & \textit{fgt} \\
\midrule
 $\lambda_{fm}=0.25$  & 59.0 & 0.09 & 64.4 & 0.06 & 50.9 & 0.08 & 53.6 & 0.08 \\
 $\lambda_{fm}=1.0$   & 61.4 & 0.07 & 65.3 & 0.05 & 53.3 & 0.07 & 54.2 & 0.07 \\
 $\lambda_{fm}=5.0$   & 62.4 & 0.06 & 66.5 & 0.04 & 54.2 & 0.06 & 56.2 & 0.06 \\
 $\lambda_{fm}=10.0$  & 62.0 & 0.06 & 66.0 & 0.03 & 53.3 & 0.06 & 53.4 &  0.07 \\
 $\lambda_{fm}=25.0$  & 62.2 & 0.06 & 63.3 & 0.03 & 52.2 & 0.07 & 51.5 &  0.05 \\
 $\lambda_{fm}=100.0$ & 58.6 & 0.04 & 55.8 & 0.02 & 46.1 & 0.04 & 42.3 &  0.03 \\
\bottomrule
\end{tabular}
}
\end{center}
\end{table*}

Effect of varying $\lambda_{fm}$ is shown in Tables.\ref{tab:taskonomy_sensitivity_analysis}, \ref{tab:split_cifar_sensitivity}. We observe that as $\lambda_{fm}$ increases, forgetting consistently falls. This is because higher values of $\lambda_{fm}$ will prevent the features from drifting. However, very high values of $\lambda_{fm}$ would regularize the models severely and impede learning. There exist an optimal $\lambda_{fm}$ that achieves the best performance while reducing forgetting.

\section{Additional results}
In Table~\ref{tab:split_cifar_255}, we provide additional results comparing CAR and the baseline algorithms on Split-CIFAR and Split-miniImageNet with memory size $m=255$.

\begin{table*}[ht]
\caption{\small\textbf{Split-CIFAR and Split-miniImageNet Results:}  Average accuracy (acc$\pm$std \%) and forgetting ($\pm$std) values are reported. Results are averaged over $5$ random seeds. The best result is highlighted in bold.}
\label{tab:split_cifar_255}
\begin{center}
\begin{tabular}{r|cc|cc}
\toprule
& \multicolumn{2}{|c||}{\sc Split-CIFAR} & \multicolumn{2}{c}{\sc Split-miniImageNet}\\
Method & \multicolumn{2}{c}{$m=255$} & \multicolumn{2}{c}{$m=255$} \\  
 & \it {acc}($\uparrow$) & \it fgt($\downarrow$) & \it acc($\uparrow$)  & \it fgt($\downarrow$)  \\
 \midrule \midrule 
 iCaRL  &  51.7\std{1.4} & 0.13\std{0.02} &  - &  - \\
 A-GEM   &  56.9\std{3.4} & 0.13\std{0.03} & 51.6\std{2.6} & 0.10\std{0.02} \\
 MER    &  57.7\std{2.6} & 0.11\std{0.01} &  49.4\std{3.4} & 0.12\std{0.02} \\
 ER     &  60.9\std{1.4} & 0.09\std{0.01} &  53.5\std{1.4} & 0.07\std{0.02}  \\
 HAL    &  62.9\std{1.5} & 0.08\std{0.01} & \bf 56.5\std{0.8} & \bf 0.06\std{0.01} \\
 \midrule
 CAR (ours)  & \bf 65.2\std{2.1} &  \bf 0.05\std{0.01} & 55.1\std{2.1} & \bf 0.06\std{0.01}  \\
\bottomrule
\end{tabular}
\end{center}
\end{table*}

\section{Experiment details}

\subsection{Taskonomy}

\subsubsection{Dataset: }
We use data from $54$ buildings of the Taskonomy dataset in our experiments. Of these buildings, $6$ randomly chosen buildings are used as training set, whiile the data from the other $48$ buildings are used for training and validation. The details of the buildings used in training and test set are provided below:

\textbf{Training set:} 
\textit{cutlerville, rockport, coffeen, tomales, pinesdale, capistrano, bettendorf, mcnary, barboursville, hacienda, uvalda, tolstoy, wyatt, haymarket, holcut, kingdom, windhorst, bonfield, monson, touhy, marksville, oyens, euharlee, glassboro, brentsville, goodview, darrtown, maben, pomaria, keiser, gloria, silva, portal, thrall, smoketown, martinville, waldenburg, lakeville, springerville, silerton, mifflintown, wainscott, superior, stokes, mayesville, wilkinsburg, archer}

\textbf{Test set:}
\textit{moberly, northgate, adairsville, cisne, marstons, woonsocket}

\textbf{Data processing:} For all pixel-wise tasks, we normalize the inputs in the range $[-1, 1]$. For depth maps, we perform a log normalization similar to \cite{zamir2018taskonomy}. We resize all inputs and output maps to $(256 \times 256)$ resolution, and perform mean normalization and random horizontal flipping as data augmentation. 

\subsubsection{Training}
A list of model architectures and optimization algorithm we use is stated in Table.~\ref{tab:exp_details_taskonomy}. Models were optimized using distributed training with a global batch size of 64. During sequential training, we train each task for $12.5 k$ iterations. Multitask model was trained for $37.5 k$ iterations.

\textbf{Experience Replay: }Since the encoder-decoder models used in Taskonomy dataset are very deep, optimizing the encoders and decoders using \eqref{eq:vanilla_er} can easily run out of memory. Therefore, instead of jointly minimizing the losses, at each step, we replay one of the old tasks with a replay probability $p_{replay}$, and train the current task with a probability $(1-p_{replay})$. This ensures that at each step, only one model is optimized thereby preventing the out-of-memory issues.

\begin{table*}
\caption{\textbf{Experiment details on Taskonomy dataset}}
\label{tab:exp_details_taskonomy}
\begin{center}
\resizebox{0.8\textwidth}{!}{
\begin{tabular}{l|l}
\toprule
    Encoder Architecture    & Resnet-50 \\
    Decoder Architecture    & 15-layer ConvNet for pixelwise tasks \\
                            & Average pooling + 1 FC layer for classification \\
    Encoder Learning rate   & 0.0001    \\
    Decoder Learning rate   & 0.0001    \\
    Optimizer               & Adam      \\
    LR Decay                & None      \\
    Batch Size              & 64        \\
    Effective number of training iters    & 12500 per task    \\
    
    \midrule
    \multicolumn{2}{c}{Experience Replay} \\
    \midrule
    Buffer filling strategy & Uniform Sampling \\
    Replay batch size       & 64 \\
    Replay probability $p_{replay}$     & 0.5 \\
    
    \midrule
    \multicolumn{2}{c}{CAR} \\
    \midrule
    Regularization function $\ell_{fm}$ & MSE loss \\
    Compression scheme      & Spatial + channel pooling \\
    Replay probability $p_{replay}$     & 0.5 \\
    
    \midrule
    \multicolumn{2}{c}{Side Tuning} \\
    \midrule
    Side Network Architecture   & 5-layer ConvNet   \\
    Side Network Learning rate  & 0.0001 \\
\bottomrule
\end{tabular}
}
\end{center}
\end{table*}

\subsection{Split-CIFAR / Split-miniImageNet}
For Split-CIFAR and Split-miniImageNet experiments, we follow the protocol used in \cite{chaudhry2018efficient} where the $100$ classes are split into $20$ tasks, yielding $5$ classes per task. So, each task is a $5$-way classification problem. Of these, we use $3$ classes for validation purposes and report performance on the remaining $17$ tasks. Each experiment is repeated for $5$ random seeds. 

\textbf{CAR: } The output of Resnet-18 encoder is a $160$-dimensional feature vector. Since the feature dimension is orders of magnitude smaller than the data dimension ($3072$ for Split-CIFAR and $21168$ for Split-miniImageNet), storing the entire feature has very small memory overhead. Hence, we do not perform any feature compression for these experiments.

\begin{table*}
\caption{\textbf{Experiment details on Split-CIFAR / Split-miniImageNet dataset}}
\label{tab:exp_details_split_cifar}
\begin{center}
\resizebox{0.8\textwidth}{!}{
\begin{tabular}{l|l}
\toprule
    Encoder Architecture    & Resnet-18 used in \cite{chaudhry2018efficient} \\
    Task head               & FC layer \\
    Encoder Learning rate   & 0.03    \\
    Task head rate          & 0.03    \\
    Optimizer               & SGD      \\
    LR Decay                & None      \\
    Batch Size              & 10        \\
    Number of training iterations    & Single epoch training    \\
    
    \midrule
    \multicolumn{2}{c}{Experience Replay} \\
    \midrule
    Buffer filling strategy & Uniform Sampling \\
    Replay batch size       & 5 \\
    
    \midrule
    \multicolumn{2}{c}{CAR} \\
    \midrule
    Regularization function $\ell_{fm}$ & MSE loss \\
    Compression scheme      & No compression used \\
    $\lambda_{fm}$          & 5.0 \\
\bottomrule
\end{tabular}
}
\end{center}
\end{table*}

\end{document}